\documentclass{article}

\usepackage{arxiv}

\usepackage[utf8]{inputenc} 
\usepackage[T1]{fontenc}    
\usepackage{hyperref}       
\usepackage{url}            
\usepackage{booktabs}       
\usepackage{amsfonts}       
\usepackage{nicefrac}       
\usepackage{microtype}      
\usepackage{lipsum}		
\usepackage{graphicx}
\usepackage{natbib}
\usepackage{doi}
\usepackage{placeins}
\usepackage{float}

\title{CD\&S Dataset: Handheld Imagery Dataset Acquired Under Field Conditions for Corn Disease Identification and Severity Estimation}

\author{ \href{https://orcid.org/0000-0002-7431-0808}{\includegraphics[scale=0.06]{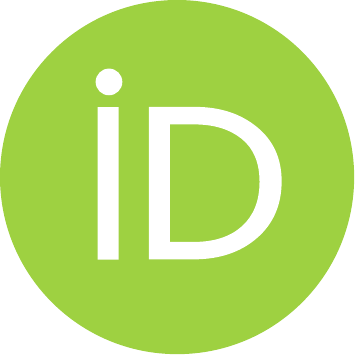}\hspace{1mm}Aanis Ahmad} \\
	Department of Electrical and Computer Engineering\\
	Purdue University\\
	West Lafayette, IN\\
	\texttt{ahmad31@purdue.edu} \\
	\And
	\href{https://orcid.org/0000-0001-8882-0510}{\includegraphics[scale=0.06]{orcid.pdf}\hspace{1mm}Dharmendra Saraswat} \\
	Department of Agricultural and Biological Engineering\\
	Purdue University\\
	West Lafayette, IN \\
	\texttt{saraswat@purdue.edu} \\
	\And
	Aly El Gamal \\
	Department of Electrical and Computer Engineering\\
	Purdue University\\
	West Lafayette, IN\\
	\texttt{elgamala@purdue.edu} \\
	\And
    Gurmukh Johal \\
	Department of Botany and Plant Pathology\\
	Purdue University\\
	West Lafayette, IN\\
	\texttt{gjohal@purdue.edu} \\
}

\renewcommand{\shorttitle}{CD\&S Dataset}

\begin{document}
\maketitle

\begin{abstract}
Accurate disease identification and its severity estimation is an important consideration for disease management. Deep learning-based solutions for disease management using imagery datasets are being increasingly explored by the research community. However, most reported studies have relied on imagery datasets that were acquired under controlled lab conditions. As a result, such models lacked the ability to identify diseases in the field. Therefore, to train a robust deep learning model for field use, an imagery dataset was created using raw images acquired under field conditions using a handheld sensor and augmented images with varying backgrounds. The Corn Disease and Severity (CD\&S) dataset consisted of 511, 524, and 562, field acquired raw images, corresponding to three common foliar corn diseases, namely Northern Leaf Blight (NLB), Gray Leaf Spot (GLS), and Northern Leaf Spot (NLS), respectively. For training disease identification models, half of the imagery data for each disease was annotated using bounding boxes and also used to generate 2343 additional images through augmentation using three different backgrounds. For severity estimation, an additional 515 raw images for NLS were acquired and categorized into severity classes ranging from 1 (resistant) to 5 (susceptible). Overall, the CD\&S dataset consisted of 4455 total images comprising of 2112 field images and 2343 augmented images.

\end{abstract}

\keywords{Plant Diseases \and Corn Diseases \and Disease Identification \and Severity Estimation \and Deep Learning \and Image Classification \and Object Detection \and Datasets}

\section{Introduction}
Corn is an important crop worldwide and diseases in corn are responsible for reducing crop yield. Northern Leaf Blight (NLB), Gray Leaf Spot (GLS), and Northern Leaf Spot (NLS), are common foliar diseases of corn found in the Midwest, United States. NLB disease is caused by the Setosphaeria turcica fungus resulting in long brown cigar shaped lesions. GLS disease is caused by Cercospora zeae-maydis and the symptoms include multiple greyish-brown and narrow rectangular lesions. Finally, NLS disease is caused by Cochliobolus carbonum resulting in multiple brown spots with circular concentric lesions. The severity of any disease in corn is dependent on varietal susceptibility, presence of the pathogen, and a favorable environmental condition (\url{https://cropprotectionnetwork.org/resources/articles/diseases/gray-leaf-spot-of-corn}). Therefore, accurate disease identification and its severity evaluation is required for implementing effective management practices.
Traditional approaches for disease identification and severity evaluation relies on manual scouting and expert knowledge. These approaches are time consuming and subjective \citep{zhang2014monitoring}. Therefore, to bring efficiency and objectivity, researchers have started exploring solutions that are based on computer vision methods. After successful application of deep learning-based image classification and object detection models across disciplines, researchers are increasingly investigating their role for disease identification and its severity evaluation by using annotated imagery datasets. However, most datasets such as PlantVillage \citep{hughes2015open} and Digipathos \citep{barbedo2018annotated}, consist of images acquired under lab conditions. As a result, models trained on such datasets failed to accurately identify diseases and their severity from images that were acquired under field conditions \citep{ferentinos2018deep, lee2020new}. 
Therefore, an annotated corn disease dataset consisting of field acquired raw images and augmented images with three varying backgrounds, was created in order to help develop robust deep learning models for disease identification and its severity estimation in the field. The field acquired raw images from the dataset were recently used \citep{ahmad2021comparison}.

\section{Materials and Methods}
\subsection{Dataset}
The CD\&S dataset consists of a total of 4455 raw images and augmented images. A total of 2112 raw images, for three common corn diseases, namely NLB, GLS, and NLS, were acquired under field conditions at Purdue University’s Agronomy Center for Research and Education (ACRE) in West Lafayette, Indiana, in July 2020. The images were acquired using a 12-megapixel camera sensor from the iPhone 11 Pro smartphone. The image mode was set to acquire images of size 3000 x 3000 pixels in order to maintain a 1:1 aspect ratio which is commonly used for the training of deep learning-based image classification and object detection models.

\begin{figure}[H]
\centering\includegraphics[width=0.9\linewidth]{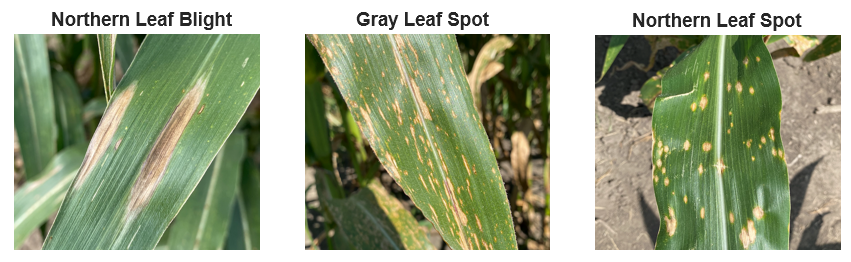}
\caption{NLB, GLS, and NLS diseases in CD\&S Dataset.}
\label{figure:1}
\end{figure}

In order to prepare the dataset for disease identification, a total of 511, 524, and 562, raw images were first acquired for the NLB, GLS, and NLS diseases, respectively (figure \ref{figure:1}). The overall dataset was then split using a 50\% - 50\% training – testing split ratio. Bounding box annotation were then created for disease lesions within the training images using the LabelImg tool (\url{https://github.com/tzutalin/labelImg}), in order to train object detection models. The images were annotated using the YOLO .txt format, which resulted in the creation of .txt files for each corresponding image. Each bounding box within an image was represented by five values (label, x-coordinate of center, y-coordinates of center, height of bounding box, width of bounding box), in each line for each annotated disease lesion.

Dataset augmentation is commonly used by increasing the dataset size and adding variations, in order to improve the generalization capability of deep learning models \citep{chao2020identification}. Therefore, the NLB, GLS, and NLS training images, that were acquired for disease identification were augmented by varying backgrounds. First, the online background removal tool (\url{https://www.remove.bg/upload}) was used for removing the field backgrounds and images were resized to 500 x 500 pixels. Background removal resulted in a total of 781 RGBA images of diseased leaves without backgrounds, where A represents the alpha channel for adjusting the transparency of pixels. A Python script was then written to place the RGBA corn disease images with no backgrounds on a white background and a black background, resulting in two additional versions of the training dataset with 781 images each. Overall, a total of 2343 augmented corn disease images were created.

For the purpose of training disease severity evaluation models, an additional 515 raw images for the NLS disease were acquired under field conditions. The total 1077 raw NLS disease images were then split into five severity classes ranging from 1 (resistant) to 5 (susceptible) with 108, 330, 297, 231, and 111 images, respectively.


The CD\&S dataset was organized into six folders (table \ref{tab:table}). Dataset\_Original consists of the raw images, for the three diseases, acquired under field conditions. The annotated images with their respective annotation files are present in Dataset\_Annotated. The augmented training images with three varying backgrounds, were stored in Dataset\_Black\_Background, Dataset\_White\_Background, and Dataset\_No\_Background. Finally, Dataset\_Severity consists of the NLS disease severity images.

\begin{table}[H]
	\caption{Folders from CD\&S Dataset}
	\centering
	\begin{tabular}{lll}
		\toprule
		\cmidrule(r){1-2}
		Folder Name     & File Type     & Link \\
		\midrule
        Dataset\_Original          & JPG files (.jpg)               & \url{https://osf.io/s6ru5/}         \\ 
        Dataset\_Annotated         & JPG \& Text files (.jpg, .txt) & \url{https://osf.io/s6ru5/}         \\ 
        Dataset\_Black\_Background & JPG files (.jpg)               & \url{https://osf.io/s6ru5/}         \\ 
        Dataset\_White\_Background & JPG files (.jpg)               & \url{https://osf.io/s6ru5/}         \\ 
        Dataset\_No\_Background    & PNG files (.png)               & \url{https://osf.io/s6ru5/}         \\ 
        Dataset\_Severity          & JPG files (.jpg)               & \url{https://osf.io/s6ru5/}         \\ 
		\bottomrule
	\end{tabular}
	\label{tab:table}
\end{table}

\bibliographystyle{model1-num-names}
\bibliography{references}  






\end{document}